\documentclass[letterpaper, 10 pt, conference]{ieeeconf}  

\IEEEoverridecommandlockouts                              

\usepackage{amsmath} 
\usepackage{amssymb} 
\usepackage{bm}
\usepackage{graphicx}
\usepackage{float}
\usepackage[linesnumbered,ruled]{algorithm2e}
\usepackage{soul,color}
\usepackage[colorlinks]{hyperref}
\usepackage[caption=false]{subfig}
\usepackage{booktabs}
\usepackage[final]{changes}

\newtheorem{prop}{Proposition}

\newcommand{\ego}{\text{ego}}

\definecolor{ao(english)}{rgb}{0.0, 0.5, 0.0}
\definecolor{alizarin}{rgb}{0.82, 0.1, 0.26}

\DeclareMathOperator*{\argmin}{\arg\!\min}

\title{\LARGE \bf
Risk-Aware Lane Selection on Highway with Dynamic Obstacles
}
\author{Sangjae Bae$^{1}$, David Isele$^{1}$, Kikuo Fujimura$^{1}$, and Scott J. Moura$^{2}$
\thanks{$^{1}$Honda Research Institute, CA, 95134 USA e-mail: {\tt\{sbae, disele, kfujimura\}@honda-ri.com}}
\thanks{$^{2}$University of California, Berkeley,
CA, 94720 USA e-mail: {\tt smoura@berkeley.edu}.}
}

\begin{document}

\maketitle

\begin{abstract}
This paper proposes a discretionary lane selection algorithm. In particular, highway driving is considered as a targeted scenario, where each lane has a different level of traffic flow. When lane-changing is discretionary, it is advised not to change lanes unless highly beneficial, e.g., reducing travel time significantly or securing higher safety. Evaluating such ``benefit'' is a challenge, along with multiple surrounding vehicles in dynamic speed and heading with uncertainty. We propose a real-time lane-selection algorithm with careful cost considerations and with a modularity in design. The algorithm is search-based optimization method that evaluates uncertain dynamic positions of other vehicles under a continuous time and space domain. For demonstration, we incorporate a state-of-the-art motion planner framework (Neural Networks integrated Model Predictive Control) under a CARLA simulation environment. 
\end{abstract}

\section{INTRODUCTION}

\begin{figure}
    \centering
    \includegraphics[width=0.8\columnwidth]{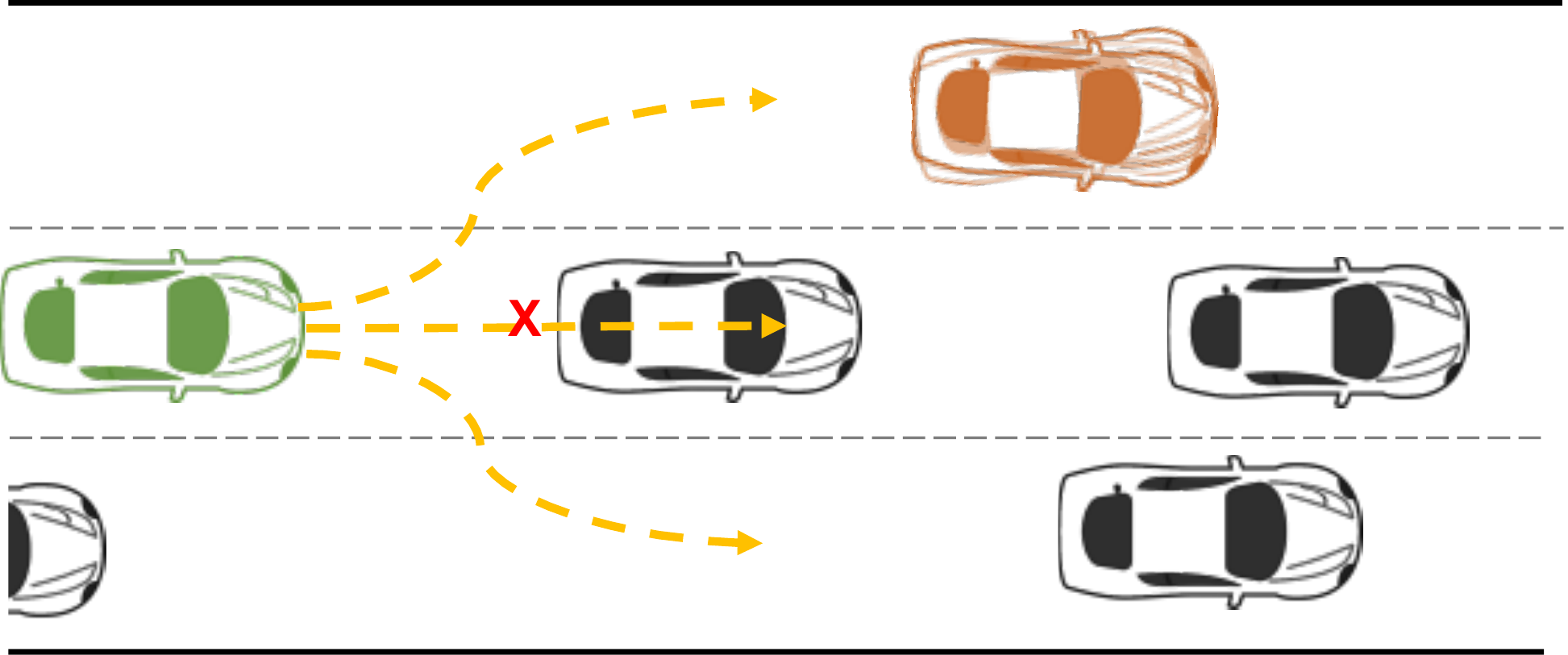}
    \caption{\textbf{Motivational example}. With a slow vehicle upfront, the ego vehicle (in \textcolor{ao(english)}{green} needs to slow down unless it changes the lane. However, the vehicle (in \textcolor{alizarin}{brown}) in the left lane is volatile and the right lane has a small inter-vehicle gap to merge into.} 
    \label{fig:motivation}
    \vspace{-10pt}
\end{figure}

Lane changing is considered one of the most risky driving behaviors \cite{liu2018risk}, as it requires multi-directional perceptions and predictions of other drivers, as well as timely decision making. An aggressive lane-changing maneuver without carefully observing other vehicles' maneuvers can cause collisions or severe effect on the safety of all surrounding vehicles. Nevertheless, \textit{if} lane-changing is executed at a right moment, lane-changing can significantly save travel time, securing a wider visibility range, and safety of driving on upcoming routes. To ground our discussion, consider being stuck behind a slow moving vehicle on the highway as illustrated in Fig.~\ref{fig:motivation}. A strategy that maintains the current lane may introduce travel delays. However, changing lanes may disrupt other traffic participants and introduce unnecessary risk. 

In fact, such lane-selection is often considered from motion planners \cite{mcnaughton2011motion, chen2019motion}, which make a lane choice and simultaneously determine a maneuver (waypoints) to merge into the selected lane. However, a critical component to many high-risk situations is the long tail of uncertain behaviors from other drivers. Because various traffic participants have different, often unknown objectives, and these objectives can conflict, there is a need to negotiate unstructured situations (for example merging in dense traffic). This negotiation requires an agent to both indicate its own intentions and interpret and respond to the intentions of others. There is an increasing body of literature related to handling these complex and interactive behaviors \cite{sadigh2016planning,isele2019interactive,schwarting2019social,bae2020cooperation}. However, a common trait of these methods is the computational complexity incurred from handling the broad uncertainty. Due to the high computation costs, and large amounts of uncertainty, these behaviors typically only plan over a short time horizon. \added[]{In fact, automated lane changes already exist in production, in for example GM SuperCruise and Tesla Autopilot \cite{olsen2018cadillac}. However, they are conservative and fail to change lanes during the highly uncertain situations examined here, and only plan over short horizons.}

Hence, this work focuses on a standalone long horizon strategic decision making process that carefully estimates benefits of lane-changing to each lane and that can be coupled with the more computationally demanding negotiation behaviors. By incorporating long horizon strategies, an agent can potentially avoid challenging interactions or configure the scene so that the agent has sufficient time or a favorable position to handle the difficult situation.



All in all, we articulate three main technical challenges to design a standalone lane-selector. \textbf{[C1]} The first challenge relates to \textit{cost formulations}, \added[]{which is fundamental in motion planning algorithms \cite{scherer2011multiple, gritsenko2015learning}}. Recall the motivational example in Fig.~\ref{fig:motivation}, there are three possible lane choices, i.e., lane-changing to left, keeping the current lane, and lane-changing to right. Each lane is faced with a unique situation, i.e., a volatile vehicle on the left, slow vehicle on the center, and small inter-vehicle gap on the right. The cost formulation, therefore, should comprehensively cover travel time, efforts to change lanes, and risks associated with limited space and volatility. \textbf{[C2]} The second challenge relates to \textit{predictions on other vehicles' position} \added[]{\cite{khandelwal2020if}}, i.e,. evaluating dynamic obstacles. Each vehicle is different and they might have a complex decision making mechanism. Therefore, the lane-selector needs to be a flexible place holder of various prediction modules, each of which may be effective in unique situations. \textbf{[C3]} The third challenge relates to \textit{computational efficiency} \added[]{\cite{barabas2017current}}. Recall that the decision making process during lane-changing must be executed in timely fashion, and therefore the computation for lane-selection needs to be prompt. Considering it as a long horizon planning with multiple lane options, a choice of problem structure and algorithm is essential in securing \textit{real-time} computation efficiency.



There exists an ample set of literature focusing on lane-selection problems. Minimizing Overall Braking Induced by Lane change (MOBIL) \cite{treiber2002minimizing} represents an initial success of automated lane selection model, which embeds Intelligent Driver Model \cite{treiber2000congested} to predict other vehicles' motions. MOBIL compares advantages and disadvantages of lane-changing to a neighboring lane, estimating an acceleration gain and loss. The success was continued by Model Predictive Control (MPC) framework \cite{schildbach2015scenario,suh2018stochastic}, with its capability of embedding predictions of other vehicles into conventional (nonlinear) optimizations that can be solved by standard nonlinear programming methods, e.g., sequential quadratic programming \cite{boggs1995sequential}. However, MPC relies on vehicle models, which are often over-simplified (or approximated) for computational reason. Their prediction models are (by structure) keenly integrated to optimization models, and hence, they are typically\footnote{Note, there exist variants that expand a range of applicable prediction modules, e.g., \cite{bae2020cooperation}.} limited to a few prediction models, such as deterministic model \cite{gray2013robust,anderson2010optimal} or hidden Markov model \cite{cesari2017scenario}. In a broader scope of applications, search-based optimization methods can be implemented in computationally efficient ways and are effective in solving nonlinear problems because they rely on forward model evaluation instead of iterative optimization.2 Examples include Dijkstra \cite{chen2003dijkstra}, Rapidly exploring Random Trees \cite{lavalle1998rapidly}, and A$^\star$\cite{hart1968formal}. In particular, A$^\star$ is generally considered one of the most computation-efficient algorithms \cite{zammit2018comparison} for its problem structure with heuristics. However, it often suffers from problem settings in continuous time and space domain and with dynamic obstacles. In short, existing methods are effective in various problem settings, however, simultaneously addressing the above challenges \textbf{[C1]}-\textbf{[C3]} remains as a research gap -- which this paper fulfills. 


The main contribution of this paper is twofold: (i) We formally design a cost function under a search-based optimization method A$^\star$, and extend the method so that dynamic obstacles are systematically evaluated. (ii) We keep the framework modular, so that any choice of prediction models (for other vehicles) and motion planners are accommodated under a certain condition. With the contribution, the challenges \textbf{[C1]}-\textbf{[C3]} are adequately addressed. We also integrate our previous motion planner design and demonstrate a complete pipeline of autonomous lane-changing.

This paper is organized in the following manner. Section~\ref{sec:algorithm} discusses the detailed cost formulations and complete search algorithm. Section~\ref{sec:simulation} presents simulation setup and reports simulation results followed by analysis. The paper concludes with a summary in Section~\ref{sec:conclusion}.

\section{LANE SELECTION ALGORITHM}\label{sec:algorithm}
\begin{figure}
    \centering
    \includegraphics[width=0.8\columnwidth]{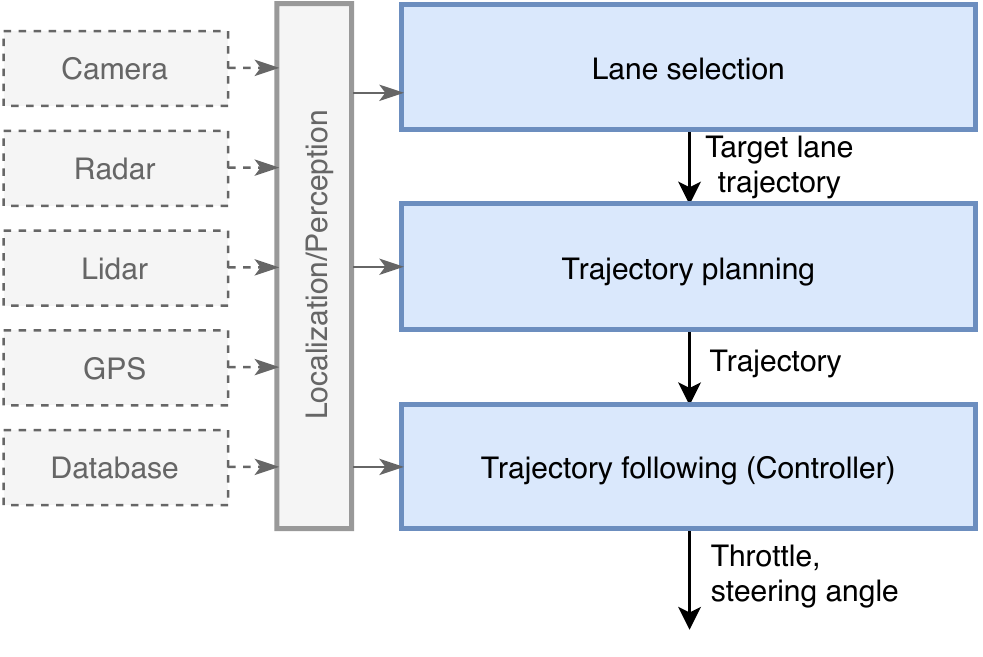}
    \caption{Pipeline of lane-changing algorithm. Given localization and perception information, the lane selection block (which is the focus of this paper) determines a target lane, the trajectory planning block determines a trajectory to the target lane, and the trajectory following (controller) block determines throttle and steering angle to actuate the maneuver.}
    \label{fig:pipeline}
\end{figure}

Figure~\ref{fig:pipeline} illustrates a complete pipeline of the lane-changing algorithm, which is composed of three layers: (i) lane selection, (ii) trajectory planning, and (iii) trajectory following. Given the geographical information of surrounding lanes, i.e., trajectory of center lines, the lane selection layer determines a target lane, considering traffic flow, risk, and travel time. The center line trajectory of the determined target lane is sent to the trajectory planning layer, which then determines a smooth position trajectory to arrive at the target lane (i.e., motion planner). Finally, the trajectory following layer computes a throttle and steering angle to actuate the maneuver (i.e., controller). 
In this paper, we focus on the first layer, i.e., the lane selection layer, and we incorporate our previous work on the trajectory planning and following algorithms \cite{bae2020cooperation} for demonstration studies. We assume the knowledge of perception and localization information, as well as the geographic information of lanes and an exit from the highway (if applicable). Noises in localization and perception are not considered in this study for simplicity.

%
%
\subsection{Graph Network}
\begin{figure}
    \centering
    \includegraphics[width=0.8\columnwidth]{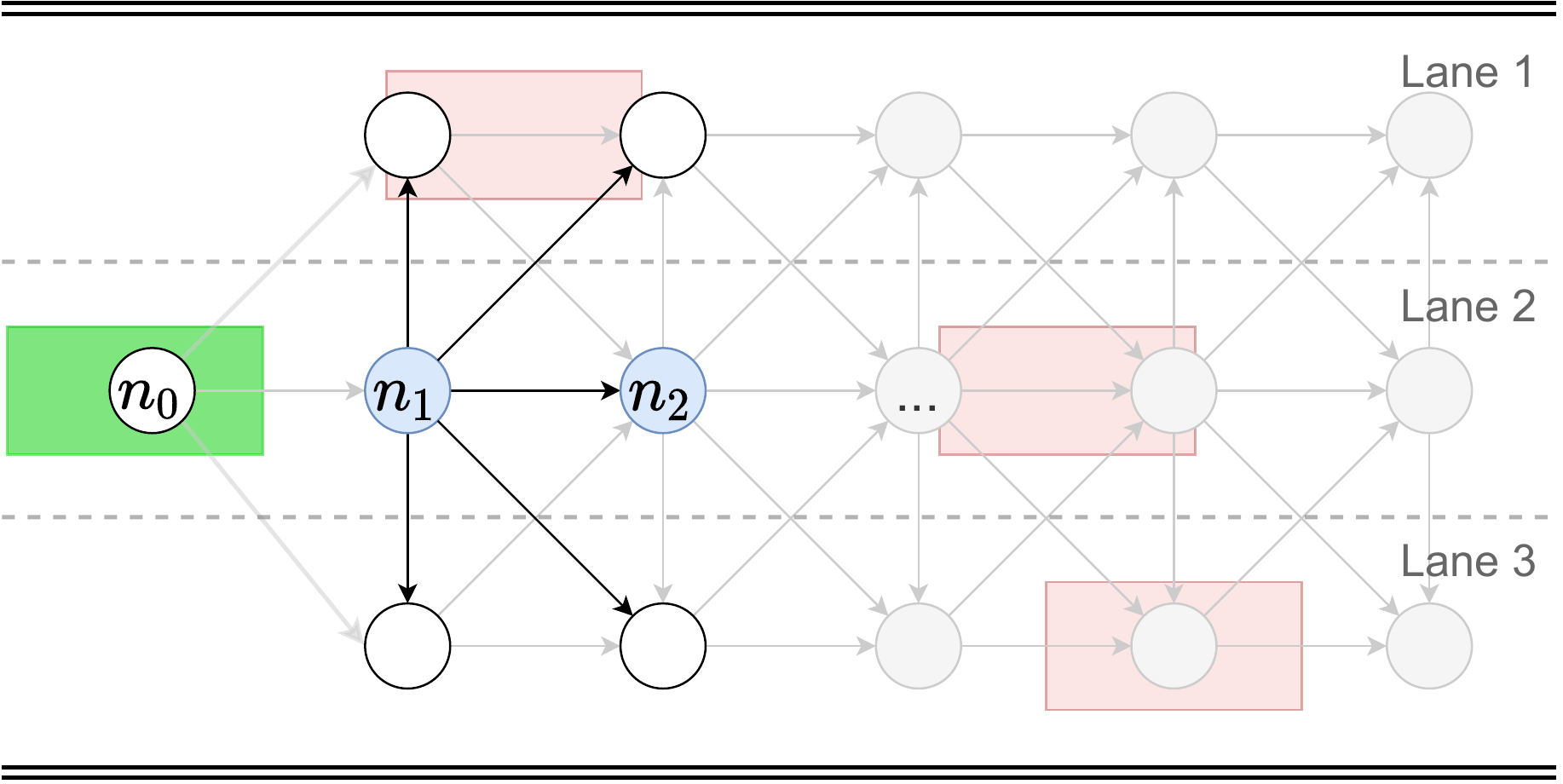}
    \caption{Graph networks for A$^\star$. Rectangle box in green indicates a measured position of the ego vehicle and rectangle boxes in red indicate measured positions of the other vehicles at the current time. At each distance step, the ego vehicle can move up to one adjacent node. For example, from node $n_0$, the ego vehicle can move to either node $n_1$, node $n_2$, or node $n_3$, which indicates lane-changing to left, lane-keeping, or lane-changing to right, respectively. Note, a perpendicular direction (to heading) is not considered, e.g., $n_1\leftrightarrow n_2$, as physically infeasible.}
    \label{fig:graph}
    \vspace{-10pt}
\end{figure}

We first generate a graph as illustrated in Fig.~\ref{fig:graph} where each node represents a position of the ego vehicle (the autonomous vehicle we control) and each edge represents a maneuver to move from one position to another. The nodes are spatially and evenly distributed at each vertical and horizontal level. The vertical level corresponds to lanes, with one row of nodes per lane, i.e., vertical step size $=$ lane width. The horizontal nodes are evenly distributed with the step size equivalent to a current speed for a fixed time step, i.e., (horizontal step size) $=$ (current speed) $\times$ (time step). Note that the ego vehicle is positioned in a discrete space, however, the other vehicles are positioned in a continuous space.

\subsubsection{Initializing Vehicles' Positions on Graph}\label{sec:init}
There are two steps of initializing positions. First, given a road angle $\psi$, each position $(x,y)$ is rotated clockwise so that the lanes are set to Eastbound. Formally, 
\begin{align}
    x^r &= x\cos\psi + y\sin\psi,\nonumber\\
    y^r &= x\sin\psi - y\cos\psi.
\end{align}
This enables indexing to be perpendicular and straightforward to analyse. Second, the rotated position $(x^r,y^r)$ is shifted and projected onto a relative space. For the ego vehicle's position $(x^r_\text{ego},y^r_\text{ego})$, the longitudinal position is set to $0$, i.e., $\Tilde{x}_\text{ego}=0$, which is a shifted position by $x^r_\text{ego}$. Similarly, vehicle $i$'s longitudinal position is also shifted by $x^r_\text{ego}$, i.e., $\Tilde{x}_i=x^r_i-x^r_\text{ego}$. Now, the lateral position of the ego vehicle is rounded to the nearest node and shifted by the leftmost lane $y^r_{\text{lane},1}$. Namely, 
\begin{equation}
    \Tilde{y}_\text{ego} = y^r_{\text{lane},j_\star}-y^r_{\text{lane},1},
\end{equation}
where
\begin{equation}
    j_\star = \argmin_{j\in\mathcal{J}}|y^r_{\text{lane},j}-y^r_\text{ego}|,
\end{equation}
with a set of lane numbers $\mathcal{J}$. The lane numbers are in the ascending order, $\{1,2,...,N_\text{lane}\}$, from the leftmost lane. This projection helps reduce dimensions, keep nodes evenly distributed with constant step size, and makes analysis straightforward. One caveat is that this could result in an offset between the original position and projected position. This offset can yield a reverting behavior while the ego vehicle is changing lane if the offset is not properly evaluated -- we will systematically address this offset in the cost formulation in the later section. 

For the other vehicles, a lateral position is projected, relative to a lane width $\ell_\text{lane}$, and shifted by the leftmost lane. Formally, for vehicle $i\in\{1,\ldots,N_\text{veh}\}$,
\begin{equation}
    \Tilde{y}_i = \frac{(y^r_i-y^r_{\text{lane},1})}{\ell_\text{lane}}.
\end{equation}
Note that the lateral position of the other vehicles is in a continuous space, and no offset exists between the original position and projected position.

\subsection{Cost Formulation}
We frame the optimization as an A$^\star$ search. The total cost $f$ at each node $n$ inherited from \added[]{its (arbitrary) parent node} $n_0$ is composed of step cost $g$ and heuristic cost $h$. Formally,
\begin{equation}
    f(n|n_0) = g(n|n_0) + h(n).\label{eq:f}
\end{equation}
The step cost $g$ represents an immediate cost of transitioning from one node to another and the heuristic cost $h$ represents an approximate optimal cost-to-go to the goal. Each cost is detailed in the following sections.

\subsubsection{Step cost}
The step cost combines four distinctive penalties: (i) control effort, (ii) travel time, (iii) risk, and (iv) switching cost. 

\vspace{3mm}
\textbf{Control effort:}
For simplicity, we consider longitudinal and lateral movement as control efforts, which are evaluated as a projection of a distance between the nodes onto the horizontal and vertical axis, respectively. From node $n_0$ to $n$, the penalty function on the control effort reads,
\begin{equation}
    g_\text{control}(n|n_0)=\lambda_\text{lng} |d(n_0,n)\cos(\theta)| + \lambda_\text{lat} |d(n_0,n)\sin(\theta)|,\label{eq:g_control}
\end{equation}
where $\lambda$ denotes weights, $d(n_0,n)$ is a Euclidean distance from node $n_0$ to node $n$, and $\theta$ is a relative angle of the transition maneuver to the lane angle. The relative angle $\theta$ is uniquely defined in two distinguished cases: whether the transition is on the same horizontal level or not. Formally,
\begin{equation}
    \theta = \begin{cases}
    \arctan{\left(\frac{\ell_{\text{lane}}}{\added[]{d(n_0,n)}}\right)}&,\quad\text{if }\Tilde{y}_{n_0}\added[]{\neq}\Tilde{y}_{n}\\
    \arctan{\left(\frac{0}{\added[]{d(n_0,n)}}\right)}=0&,\quad\text{if }\Tilde{y}_{n_0}\added[]{=}\Tilde{y}_{n}
    \end{cases},
\end{equation}
where $\Tilde{y}_{n_0}$ and $\Tilde{y}_{n}$ denotes a projected lateral position of node $n_0$ and node $n$. Recall that lane-changing behavior is risky and therefore the lateral control is more significantly penalized than the longitudinal control. Mathematically, we impose a high weight on lateral control compared to longitudinal control, i.e., $\lambda_{\text{lat}} \gg \lambda_{\text{lng}}$.

\vspace{3mm}
\textbf{Travel time:} A travel time saving is often a main motivation of lane-changing, and hence an essential part in the cost function. Given a current speed $v$, the transition time in seconds from node $n_0$ to node $n$ is written
\begin{equation}
    t(n_0,n) = \frac{d(n_0,n)}{v}.\label{eq:t}
\end{equation}
Since the distance step is not identical between longitudinal and lateral directions, the distance $d(n_0,n)$ depends on the two cases of the lateral position of the nodes: (i) if node $n$ is in the same lateral position of node $n_0$, i.e., $\Tilde{y}_{n_0}=\Tilde{y}_{n}$, (ii) if not, i.e., $\Tilde{y}_{n_0}\neq\Tilde{y}_{n}$. 
Hence, from node $n_0$ to node $n$ the penalty function on the travel time reads,
\begin{align}
    g_{\text{time}}(n|n_o)&=\lambda_{\text{time}}t(n_0,n)\\
    &=\begin{cases}
    \lambda_\text{time}&,\quad\text{if }\Tilde{y}_{n_0}=\Tilde{y}_{n}\\
    \lambda_\text{time}\frac{\sqrt{(\ell_\text{lane})^2+(v)^2}}{v^2}&,\quad\text{if }\Tilde{y}_{n_0}\neq\Tilde{y}_{n}
    \end{cases}.
\end{align}
Now, we evaluate additional travel time due to a slow vehicle upfront. If the front vehicle speed $v_f$ is lower than the ego vehicle's, we suppose that the ego vehicle must follow the speed of a front vehicle, and the reduced speed is penalized in the form of additional travel time $t_\text{add}(n_0,n)$. Formally, 
\begin{equation}
    t_\text{add}(n_0,n) = \frac{\added[]{(v-v_f)_+}}{d(n_0,n)},
\end{equation}
\added[]{where $(\cdot)_+$ indicates a bounded positive value, i.e., $\max(0, \cdot)$}. 
The complete cost function of the travel time reads,
\begin{equation}
    g_{\text{time}}(n|n_o)=\lambda_{\text{time}}\bigg( t(n_0,n) + t_\text{add}(n_0,n)\bigg).\label{eq:g_time}
\end{equation}

\vspace{3mm}
\textbf{Risk:} To properly balance out the travel time savings against driving risk, the penalty on the risk needs formal evaluation. We conjecture that the risk increases in two cases. First, when a physical distance to a neighboring vehicle is short (adjacency risk). Second, when a neighboring vehicle vacillates (uncertainty risk). 

The adjacency risk is simply evaluated by an inverse Euclidean distance, 
which leads the risk to increase exponentially as the distance decreases. At node $n$, the adjacency risk suffices
\begin{equation}
    r_\text{adj}(n) = \lambda_\text{adj}\sum_{i\in\mathcal{I}_n}\bigg( d^2(n,\text{Veh}_i(t_n))\bigg)^{-1},\label{eq:r_adj}
\end{equation}
 where $\mathcal{I}_n$ denotes a set of vehicle indices on the same lane with node $n$, $\text{Veh}_i(t_n)$ denotes an estimated position of vehicle $i$ at time $t_n$, and $t_n$ indicates travel time from the current measured position to node $n$. This adjacency risk essentially assesses an overall risk in each lane, which enables the ego vehicle to choose a lane with less traffic density. 

The uncertainty risk measures how volatile the other vehicles are. For instance, if one vehicle often has a hard break and hard acceleration, the lane (only) with the volatile vehicle is not recommended to drive on. In this case, another lane with multiple less-volatile vehicles could be a better choice. We gauge such uncertainty risk by entropy from the information theory \cite{gray2011entropy} with an empirical distribution of accelerations. The empirical distribution is constructed based on real-time observations (to adapt to real-time changes) and we apply Bayesian inference \cite{box2011bayesian} to update a distribution with new observations. Hence, the uncertainty risk reads
\begin{equation}
    r_\text{uncert}(n) = \lambda_\text{uncert} \sum_{i\in\mathcal{I}_n} \frac{\text{H}(A_i)}{d^2(n,\text{Veh}_i(t_n))},
\end{equation}
where $\text{H}$ denotes an entropy function and $A_i$ denotes a discrete random variable for acceleration of vehicle $i$. 

The risk cost is then the sum of the adjacency risk and uncertainty risk, i.e.,
\begin{equation}
    g_\text{risk}(n)=r_\text{adj}(n)+r_\text{uncert}(n).\label{eq:g_risk}
\end{equation}

\vspace{3mm}
\textbf{Switching cost:} We suppose that consecutive lane-changing is not recommended, unless necessary for safety reasons. That is, if a target lane is updated while changing-lane, it can result in a waving maneuver, which is not recommended for both drive comfort and safety. The switching cost is written
\begin{equation}
    g_\text{switch}(n) = \Lambda_\text{switch}(\Delta)d(\Tilde{y}_n,y_{\star,\text{prev}}),\label{eq:g_switch}
\end{equation}
where $\Lambda_\text{switch}(\Delta)$ is a dynamic weight as a function of divergence of the initial ego position from the previous target lane, $\Delta$, and \added[]{$y_{\star,\text{prev}}$ is the lateral position of the previous target lane}. This adaptive weight is a key to prevent a waving behavior (returning back to source in the middle of lane-changing) resulting from the offset between the original and projected position of the ego vehicle, as discussed in Section~\ref{sec:init}. The dynamic weight increases if the ego vehicle is lane-changing, i.e., $|\Delta| > 0$, and it decreases if the ego vehicle finished lane-changing, i.e., $|\Delta| \approx 0$. Formally,
\begin{equation}
    \Lambda_\text{switch}(\Delta) = \lambda_\text{switch}\frac{|\Delta|}{\ell_\text{lane}}.
\end{equation}

\vspace{3mm}
\textbf{Complete step cost formulation:} The complete step cost is the sum of each penalty function, i.e., for node $n$ inherited from $n_0$, \added[]{plus a Euclidean distance to the goal position},
\begin{align}
    g&(n|n_0;y_{\star,\text{prev}})\\&= g_\text{control}(n|n_0) + g_\text{time}(n|n_0)+ g_\text{risk}(n|n_0) \nonumber\\&+ g_\text{switch}(n|n_0;y_{\star,\text{prev}}) \added[]{+\lambda_{\text{goal}}d(n,\text{goal})}\label{eq:g}\\
    &= \eqref{eq:g_control} + \eqref{eq:g_time} +\eqref{eq:g_risk} + \eqref{eq:g_switch} \added[]{+\lambda_{\text{goal}}d(n,\text{goal})}.\nonumber
\end{align}
\added[]{The distance to the goal, $\lambda_{\text{goal}}d(n,\text{goal})$, is added for the admissibility, which is detailed in Section \ref{sec:heuristic}.} Note that hard constraints are absent. In particular, a collision is treated as an additional travel cost and risk. The absence of hard constraints ensures the existence of a feasible solution from any search. In addition to providing solution guarantees, this formulation positions the vehicle to only engage in the provably lowest cost interactions necessary. 

\subsubsection{Heuristic cost}\label{sec:heuristic}
Given the geographical information, the heuristic cost $h$ approximately measures the \textit{cost-to-go} to the goal. Formally, for node $n$,
\begin{equation}
    h(n) = \lambda_\text{goal} d(n, \text{goal}),\label{eq:h}
\end{equation}
where the weight $\lambda_\text{goal}$ adaptively increases as the ego vehicle gets closer to the goal point (highway exit), i.e., $\lambda_\text{goal} \propto \frac{1}{d(\text{ego}, \text{goal})}$. \added[]{The heuristic cost is admissible, i.e., a lower bound of the optimal cost, since the step cost is strictly positive and lower bounded by the heuristic cost.}

%
\subsection{Extended A$^{\star}$ with Transition Time}
\begin{figure}
    \centering
    \includegraphics[width=0.3\columnwidth]{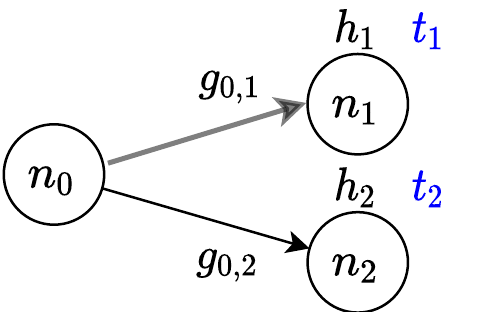}
    \caption{Illustration of Extended A$^\star$ variables configuration. At each transition, the transition time $t_n$ from the current measured position is calculated and recorded.}
    \label{fig:transition_time}
    \vspace{-10pt}
\end{figure}

A vanilla A$^{\star}$ algorithm is based on the static time and space. However, autonomous driving on the road should be keenly planned in the continuous time and space, due to the existence of dynamic obstacles (other vehicles and/or pedestrians). Therefore, as illustrated in Fig.~\ref{fig:transition_time}, we introduce an additional variable $t_n$ at each node $n$ that represents a transition time from the initial position of the ego vehicle at the current time measurement $t_0$. This transition time enables the estimation of future positions of surrounding vehicles over the planning time horizon.

\subsection{Estimation of Other Vehicles' position}
It is essential to estimate the positions of the other vehicles at each space step (i.e., at each transition of the ego vehicle from one node to another), especially for precise evaluation of additional travel time cost \eqref{eq:g_time} and risk cost \eqref{eq:g_risk}. We highlight that the proposed framework is flexible with any prediction module that outputs trajectory given observations regardless of its model. That is, any prediction module $\Phi$ is applicable that suffices
\begin{equation}
    \text{Veh}_i(t_n) = \Phi(\text{Veh}_i(t_{-N}),...,\text{Veh}_i(t_{-1}),\text{Veh}_i(t_0),t_n),\label{eq:pred_module}
\end{equation}
where $\text{Veh}_i(\cdot)$ denotes a position of vehicle $i$ at time $(\cdot)$, $t_0$ is a current time in measurement, $t_{-1}$ is a previous time step, and $t_{-N}$ is a previous $N$ time step. Recall, $t_n$ is a travel time to node $n$ from the initial position $\text{Veh}_i(t_0)$. 

An example prediction model with constant velocity \eqref{eq:pred_module} is
\begin{align}
    \text{Veh}_i(t_n) &= (x_i,y_i)|_{t_n}\nonumber\\
                      &= (x_{i0}+v_i\cos\Psi_i t_n, y_{i0}+v_i\sin\Psi_i t_n),\label{eq:pred_const_vel}
\end{align}
where $(x_{i0},y_{i0})$ is an initial position at $t_0$ and $\Psi_i$ is an inertial heading angle of vehicle $i$ relative to the road angle $\psi$. In demonstration studies, we will apply \eqref{eq:pred_const_vel} for its simplicity, however, again, more advanced prediction modules can be applied, such as Markov chain methods  \cite{okamoto2017similarity}. Also, cooperative behaviors can be evaluated under this framework, using a prediction module that considers interactions between agents, such as social generative adversarial networks \cite{gupta2018social} or graphical networks with intention reasoning \cite{choi2019drogon}.

%
\subsection{Termination Criteria with Surrogate Goals}\label{sec:termination_criteria}
Simply stated, the A$^\star$ algorithm terminates when a path reaches the goal position, i.e., $n = n_\text{goal}$. However, in the case of driving on the highway, if an exit is positioned a few thousand meters away, the goal may not be within the search horizon. There could be multiple techniques to address this issue, such as projecting a goal to the nearest node within a search horizon. In this work, we relieve the termination criteria, by having a surrogate goal at each longitudinal end of lanes. The algorithm stops when a path reaches at any end with a minimum cost, i.e., $n \in \{n'_{\text{goal},1},\ldots,n'_{\text{goal},N_{\text{lane}}}\}$, where the superscript $'$ indicates a surrogate goal node. 

\subsection{Complete Algorithm: Extended-A$^\star$}
In brief, given a current node and goal node, the \textit{Extended-A}$^\star$ finds a path by investigating nodes prioritized by low cost until a search reaches a surrogate goal node. With variations specifically for the lane-selection problem, we detail each step of search as follows.

\small{
\vspace{3mm}
\textbf{Input:}
\begin{itemize}
    \item Current state of the ego vehicle, $(x_\ego, y_\ego, v_\ego)$,
    \item Current state of vehicle $i$ $(x_i, y_i, v_i, \Psi_i)$ for $i\in\{1,\ldots,N_\text{veh}\}$,
    \item Lane $j$'s center-line trajectory $L_j$ for $j\in\{1,\ldots,N_\text{lane}\}$,
    \item Goal position $(x_\text{goal},y_\text{goal})$.
\end{itemize}

\vspace{1mm}
\textbf{Output:}
\begin{itemize}
    \item Sequence of nodes (path).
\end{itemize}

\vspace{1mm}
\textbf{Algorithm:}
\begin{enumerate}
    \item Initialize a graph network $\mathcal{G}$ by projecting vehicles onto a relative space following Section~\ref{sec:init}.
    \item Add a starting node to the open list $\ell_\text{open}$.
    \item Repeat
    \begin{enumerate}
        \item For each adjacent node\footnote{A total of three, lane-changing to left, keeping lane, and lane-changing to right.} $n$, 
        \begin{itemize}
            \item Ignore the node if $n\in\ell_\text{closed}$. Continue to the next step, otherwise.
            \item If $n\notin\ell_\text{open}$, 
            \begin{itemize}
                \item Add the node to $\ell_\text{open}$.
                \item Set the current node $n_0$ to the parent node $n_p$. 
                \item Compute the travel time $t(n|n_0)$ in \eqref{eq:t}.
                \item Compute the total travel time $t_n=t_{n_0}+t(n|n_0)$.
                \item Predict positions of other vehicles with \eqref{eq:pred_module}, over $t_n$.
                \item Compute $f$ in \eqref{eq:f}, $g$ in \eqref{eq:g}, and $h$ in \eqref{eq:h}, and record $t_n$.
            \end{itemize} 
            \item If $n\in\ell_\text{open}$, update $n_p$, $f$, $g$, $h$, and $t_n$ with a lower $g$.
        \end{itemize}
        \item Find a node with lowest $f$ in \eqref{eq:f} and add the node to the closed list $\ell_\text{closed}$. Set the node to the current node $n_0$.
    \end{enumerate}
\end{enumerate}

\vspace{1mm}
\textbf{Stop:}
\begin{itemize}
    \item When any of the surrogate goals is in $\ell_\text{close}$, i.e., if $n'\in \ell_\text{close}$ where $n'\in \{n'_{\text{goal},1},\ldots,n'_{\text{goal},N_{\text{lane}}}\}$.
\end{itemize}
}

\normalsize
\vspace{3mm}
Recall that any of the surrogate goals is reachable, and hence there is no stopping criteria related to failing to find an admissible path. \added[]{Also recall, the heuristic cost is admissible and thus the optimality of solutions is guaranteed (even with multiple surrogate goals, proved in Appendix).} 

\section{SIMULATIONS}\label{sec:simulation}

%

%
\subsection{Implementation Setup}
\begin{figure}
    \centering
    \includegraphics[width=0.7\columnwidth]{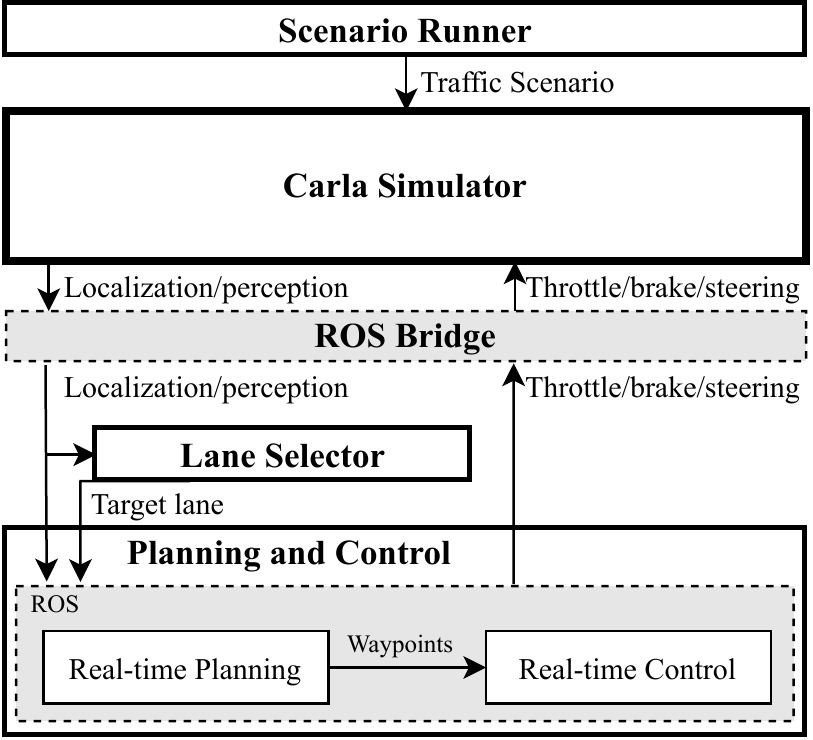}
    \caption{Implementation setup. ROS (Robot Operating System) bridge, lane selector, planner, and controller are represented by each ROS node. The scenario runner initializes a traffic scenario under the Carla simulator. Radars, Lidars, and cameras are assumed installed on the ego vehicle and no errors are imposed in localization and perception.}
    \label{fig:implementation_setup}
\end{figure}

As shown in Fig.~\ref{fig:implementation_setup}, the implementation setup is composed of three main components: (i) CARLA simulator \cite{dosovitskiy2017carla}, (ii) Lane selector, (iii) Planner and Controller \cite{bae2020cooperation}. 
The lane selector component represents the proposed method in this paper and we integrate the planner and controller design in the previous work \cite{bae2020cooperation}. We run the simulations on Ubuntu 16.04 LTS (Intel Xeon CPU ES-2640 v4 @2.40GHz x 20, GeForce GTX TITAN). For each search, the proposed method takes 0.005 [s] on average, which indicates a strong potential as an online controller.
\begin{figure}
    \centering
    \includegraphics[width=0.55\columnwidth, trim=7.6cm 2.8cm 7.6cm 1cm, clip]{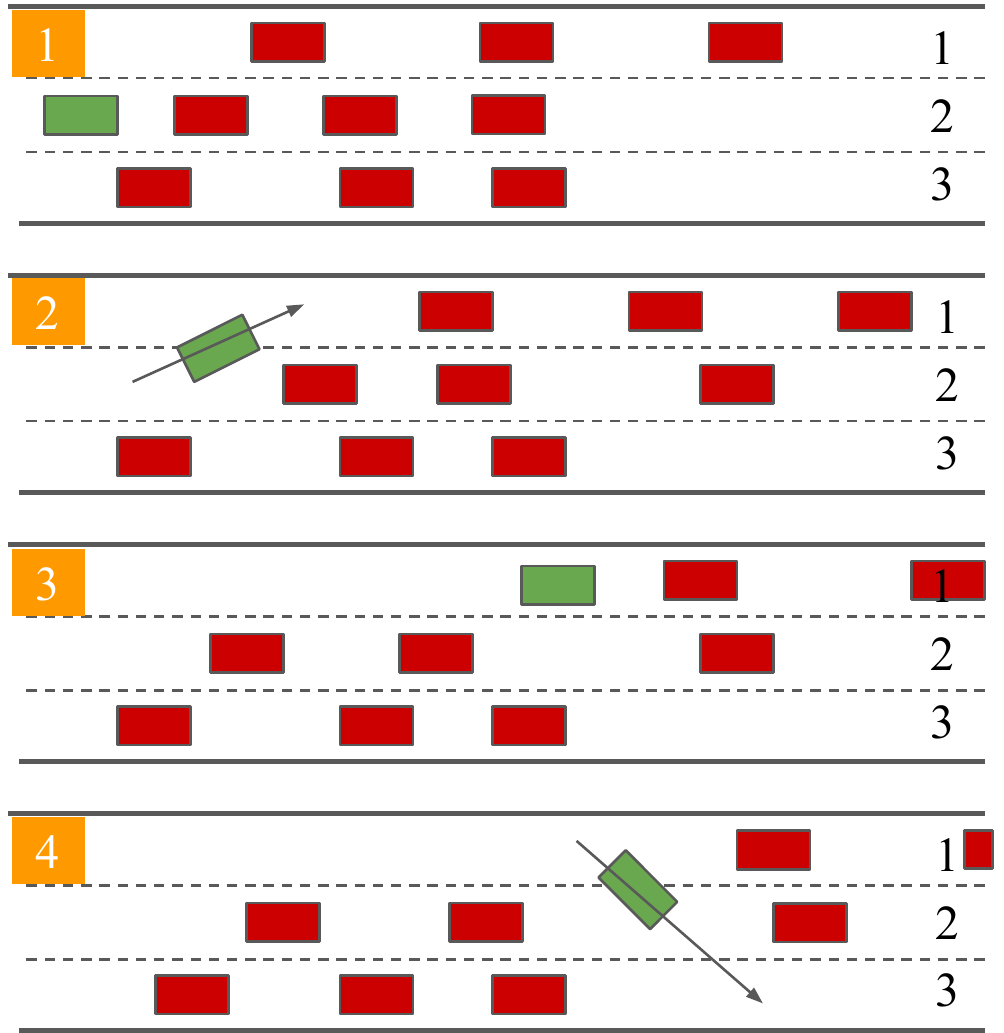}
    \caption{Driving scenario on the highway and expected behaviors of the ego vehicle (in \textcolor{ao(english)}{green}) conditioned to other vehicles (in \textcolor{red}{red}). Scenes (from 1 to 4) are in order of time, where Scene 1 illustrates initial positions of the vehicles.}
    \label{fig:scenario}
    \vspace{-10pt}
\end{figure}

\subsection{Simulation Overview}
Figure~\ref{fig:scenario} depicts a driving scenario on a segment of highway. Specifically, Scene 1 and 3 represent main decision making situations, and Scene 2 and 4 illustrate expected behaviors from Scene 1 and 3, respectively. The ego vehicle is initialized to follow a slow vehicle as in Scene 1. Given that situation, the ego vehicle is expected to change lanes to the left lane (lane 1) as a delay is foreseen in the current lane, as illustrated in Scene 2. Similarly, in Scene 3, the ego vehicle is again stuck behind a slow vehicle in lane 1. However, in this case, the right lane (lane 2) is not a good target as another vehicle ahead is also driving slow. Therefore as in Scene 4, the ego vehicle is expected to drive to lane 3, where the longest headway is achieved. 

Each lane is configured so that lane 1 has the highest throughput and lane 3 has the lowest throughput, as tabulated in Table~\ref{table:lane_config}. The desired speed of the ego vehicle is set higher (15 [m/s]) than the average speed in any lanes, and thus lane-changing is encouraged. Note that the highway exit is set to be arbitrarily far from the testing route and the route is a total of 230 [m] long\footnote{This is a fairly short route for demonstration and validation purpose.}. \added[]{The parameters in \eqref{eq:g} are set to: $\lambda_\text{lng}=1, \lambda_\text{lat}=15,\lambda_\text{adj}=\lambda_\text{uncert}=6, \lambda_\text{time}=20$, and $\lambda_\text{switch} = 7.$} The other vehicles are modeled with the Intelligent Driver Model \cite{kesting2010enhanced} with parameters set as in \cite{bae2020cooperation}. 

For comparative analysis, we consider a benchmark lane-selection model, Minimizing Overall Braking Induced by Lane-change (MOBIL) \cite{treiber2002minimizing}, as a baseline. Concisely described, MOBIL compares an advantage of changing lane to an adjacent lane against a disadvantage. The advantage is measured by an increase in accelerations of the ego vehicle and the disadvantage is measured by a decrease in accelerations of surrounding vehicles. We also compare it with ``no lane-change'' model, i.e., staying in the current lane. The performance of the proposed method is mainly evaluated by travel time and visibility range (measured by headway). 

\begin{table}[]
\centering
\begin{tabular}{@{}llll@{}}
\toprule
 & \textbf{Lane 1} & \textbf{Lane 2} & \textbf{Lane3} \\ \midrule
Average speed [m/s] & 8 & 5 & 1 \\
Density [veh/100m] & 3 & 3 & 4 \\
Average headway [m] & 30 & 25 & 20 \\ \bottomrule
\end{tabular}
\caption{Traffic configuration of Each Lane}
\label{table:lane_config}\vspace{-20pt}
\end{table}

\begin{figure}[t]
    \centering
    \includegraphics[width=0.9\columnwidth]{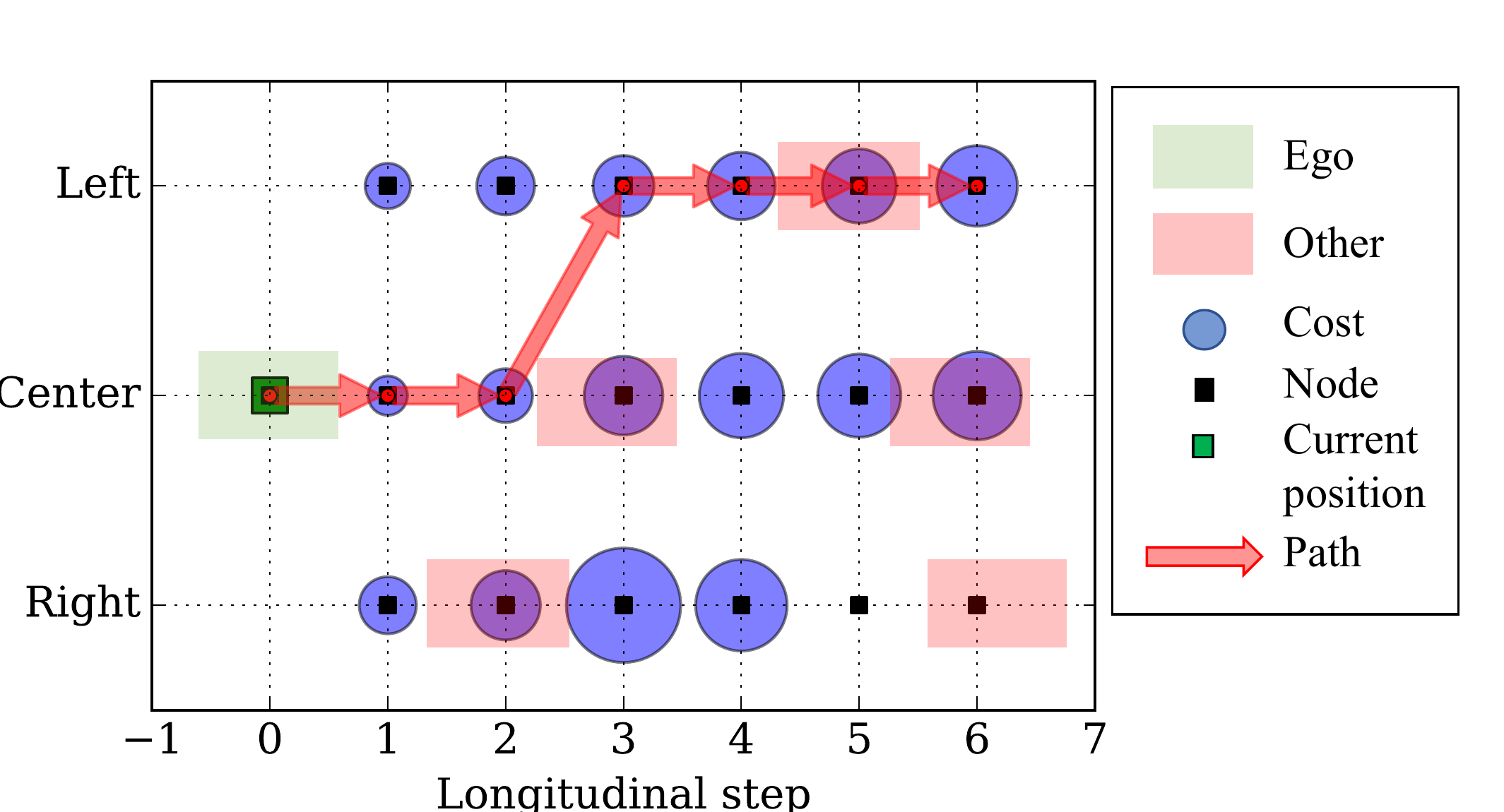}
    \caption{\textit{Extended-A}$^\star$ search instance for scene 1 in Fig.~\ref{fig:scenario}. The rectangles (in \textcolor{ao(english)}{green} and \textcolor{red}{red}) indicate initial positions (of the ego vehicle and other vehicles, respectively). The size of the blue circles (cost) indicates a value of the cost, i.e., bigger circle, higher cost.}
    \label{fig:cost}
    \vspace{-10pt}
\end{figure}

\begin{figure*}
    \centering
    \includegraphics[width=0.9\textwidth,trim=1cm 3.5cm 1cm 2.5cm,clip]{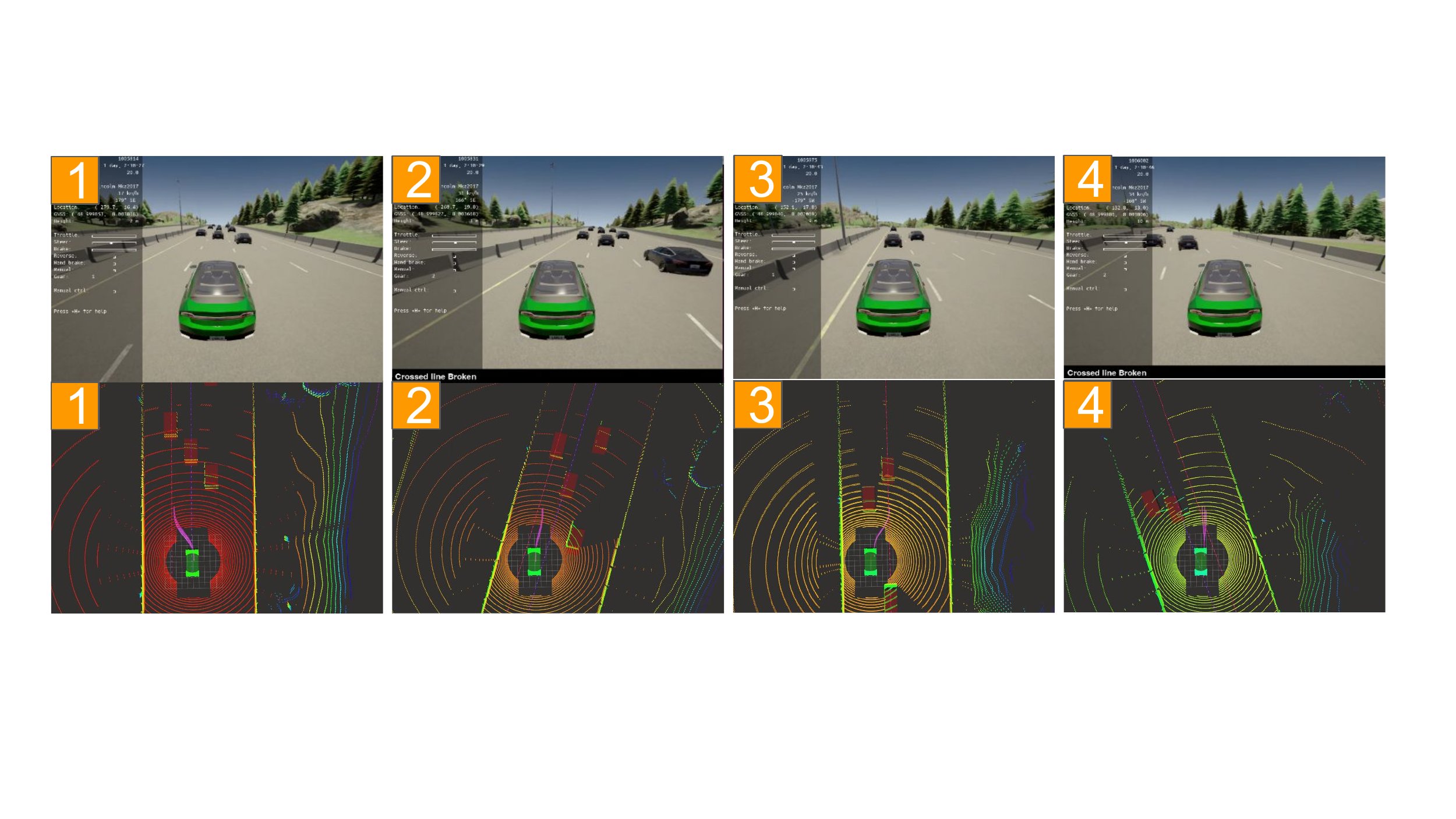}
    \caption{Lane-selection algorithm using EA$^\star$ in Carla simulator. Each scene corresponds to the scene in Fig.~\ref{fig:scenario}.}
    \label{fig:carla-demo}
    \vspace{-10pt}
\end{figure*}
\subsection{Results}


Figure~\ref{fig:cost} demonstrates an optimal path obtained by the extended A$^\star$ for the decision making situation in Scene 1 (in Fig.~\ref{fig:scenario}). The vehicles in the right lane are close to the ego vehicle and they are moving slow. Hence, a huge cost (particularly for travel time \eqref{eq:g_time} and risk \eqref{eq:g_risk}) is evaluated for lane changing to the right lane. Similarly, vehicles in the current (center) lane are driving slower, which results in high cost in travel time and risk. Eventually, the algorithm finds a path to the left lane, sacrificing a transition cost \eqref{eq:g_control}.

Similarly, Fig.~\ref{fig:carla-demo} demonstrates the ego vehicle's maneuver for each scene (Fig.~\ref{fig:scenario}), simulated by Carla. We highlight that in each decision making situation (Scene 1 and 3), the proposed algorithm properly determines a path aligned with the expected behaviors (Scene 2 and 4). We also observed that the algorithm decides to change lane sooner than later if lane change is determined. These behaviors result from risk-aware characteristics of the algorithm, which tries to keep distance to other vehicles while reducing travel time. This risk-aware decisions can be tuned by the penalty weight $\lambda_\text{adj}$ in \eqref{eq:r_adj}. That is, if we set a small value for $\lambda_\text{adj}$, the algorithm would defer lane changes until headway to the front vehicle in the current lane decreases -- this could be an interesting sensitivity analysis, however, we omit it due to the space limitation.

\begin{figure*}
    \centering
    \includegraphics[width=0.6\columnwidth]{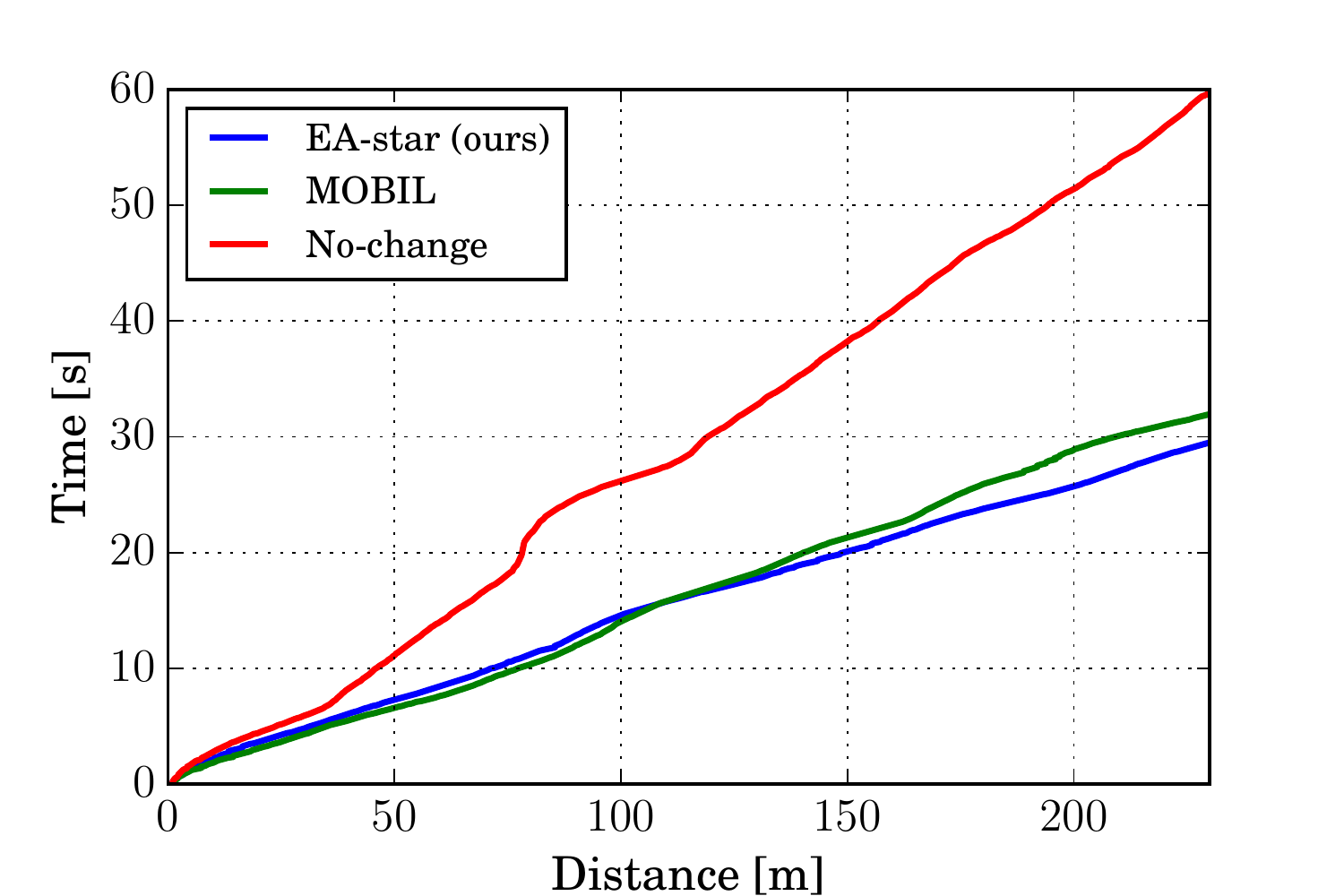}
    \includegraphics[width=0.6\columnwidth]{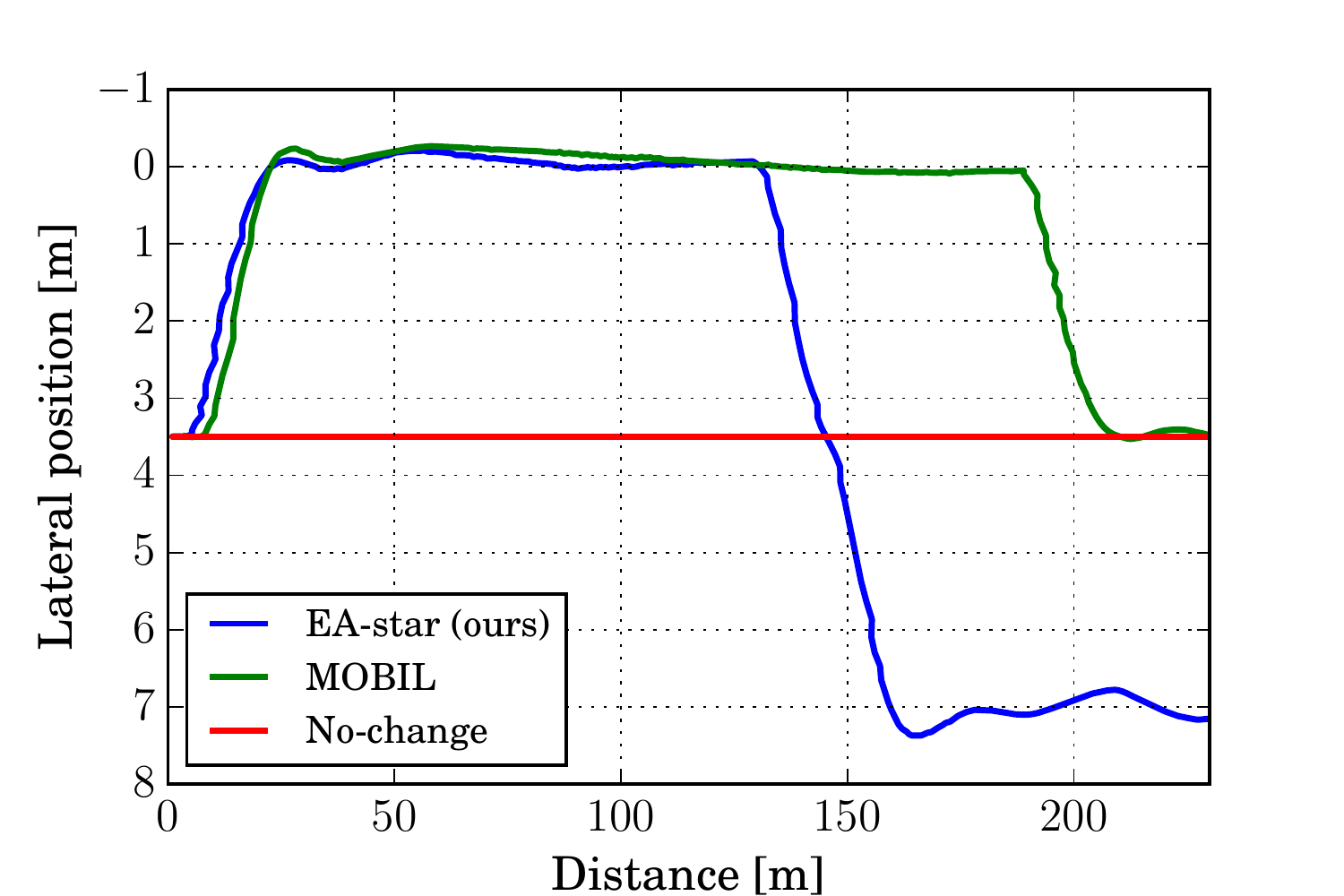}
    \includegraphics[width=0.6\columnwidth]{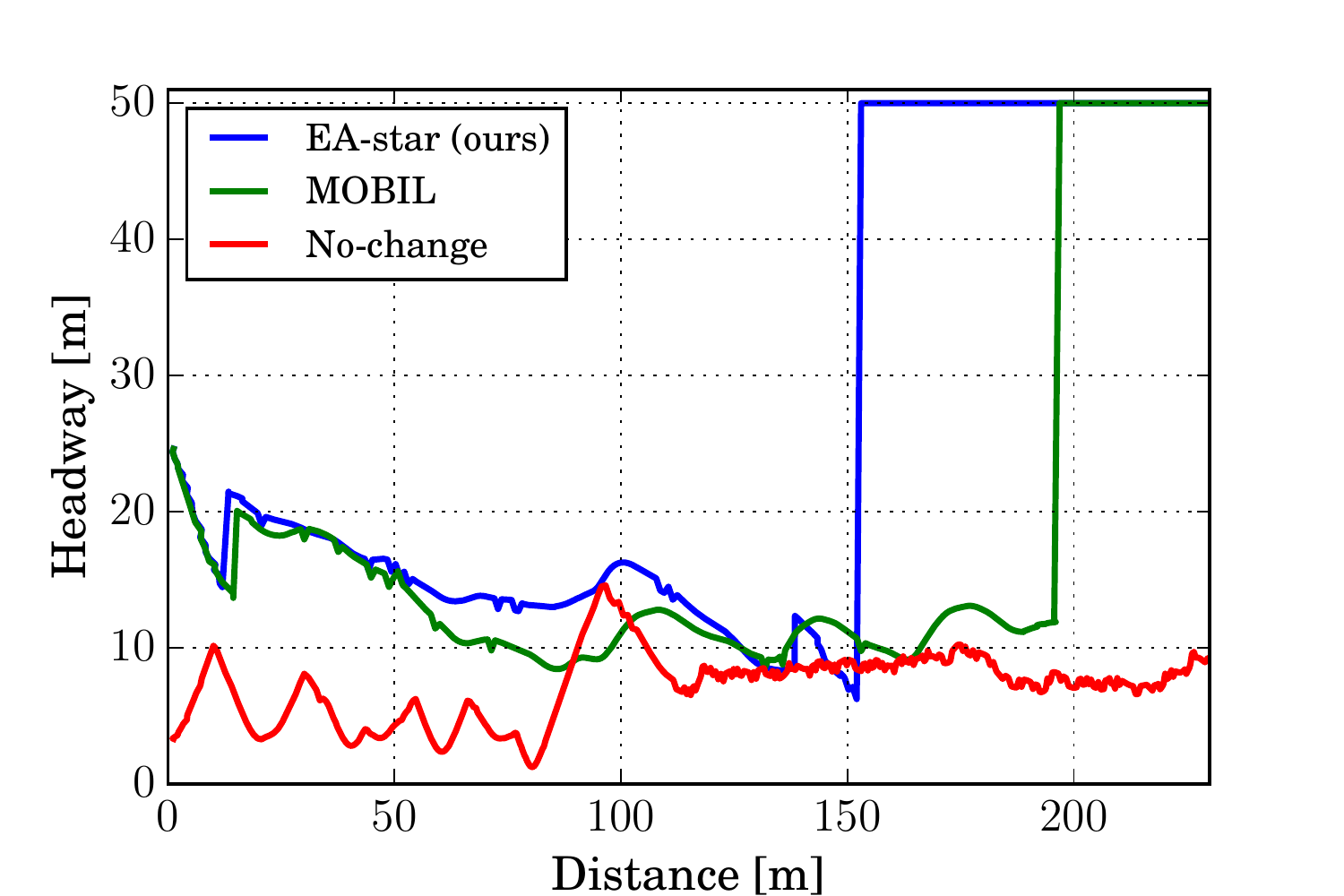}
    \caption{\added[]{LEFT: Travel time comparison. CENTER: Lane choice (lateral position) comparison. The value of 0 aligns with lane 1 (leftmost), value of 3.5 aligns with lane 2 (center), and value of 7 aligns with lane 3 (rightmost). RIGHT: Headway comparison. When no front vehicle exists, the headway is set to the detection range (50 [m]).}}
    \label{fig:trajectory}
    \vspace{-10pt}
\end{figure*}

Figure~\ref{fig:trajectory} presents trajectories for travel time, lateral position (lane choice), and headway, respectively, over the route. Those profiles are compared among EA$^\star$, MOBIL, and No-change. Most of all, the proposed algorithm using EA$^\star$ outperforms the other methods in terms of travel time, which is a main motivation for lane-changing; EA$^\star$ has 7.63\% of travel time savings compared to MOBIL and 50.53\% compared to No-change. The improvement of EA$^\star$ against MOBIL is driven by its capability of exploiting a wider range of lanes as opposed to MOBIL which only investigates adjacent lanes (either left or right). Particularly in Scene 3, MOBIL can only evaluate the advantage of lane-changing to the right lane (lane 2) which also has a slow-moving vehicle ahead. As a consequence, MOBIL ends up staying in the current lane (lane 1) albeit the second next lane (lane 3) is empty, hence being myopic. Such lane choice is clearly observed from \added[]{the center plot in }Fig.~\ref{fig:trajectory}. At around 130 [m] of travel distance, EA$^\star$ finds a path to change lane from lane 1 to lane 3, while MOBIL stays in lane 1 for another hundred meters. Meantime, No-change stays in lane 2, which is the lane the ego vehicle is initially positioned at. \added[]{The right plot in Fig.}~\ref{fig:trajectory} indicates how the headway (to a front vehicle) varies over travel distance. Over the travel distance, the headway with EA$^\star$ is generally comparable to that with MOBIL. However, EA$^\star$ secures a high headway sooner than  MOBIL, by lane-changing to lane 3 in Scene 3, which yields 47.7\% increase in the average headway against MOBIL (277.96\% increase against No-change).

\begin{figure}
    \centering
    \includegraphics[width=1\columnwidth]{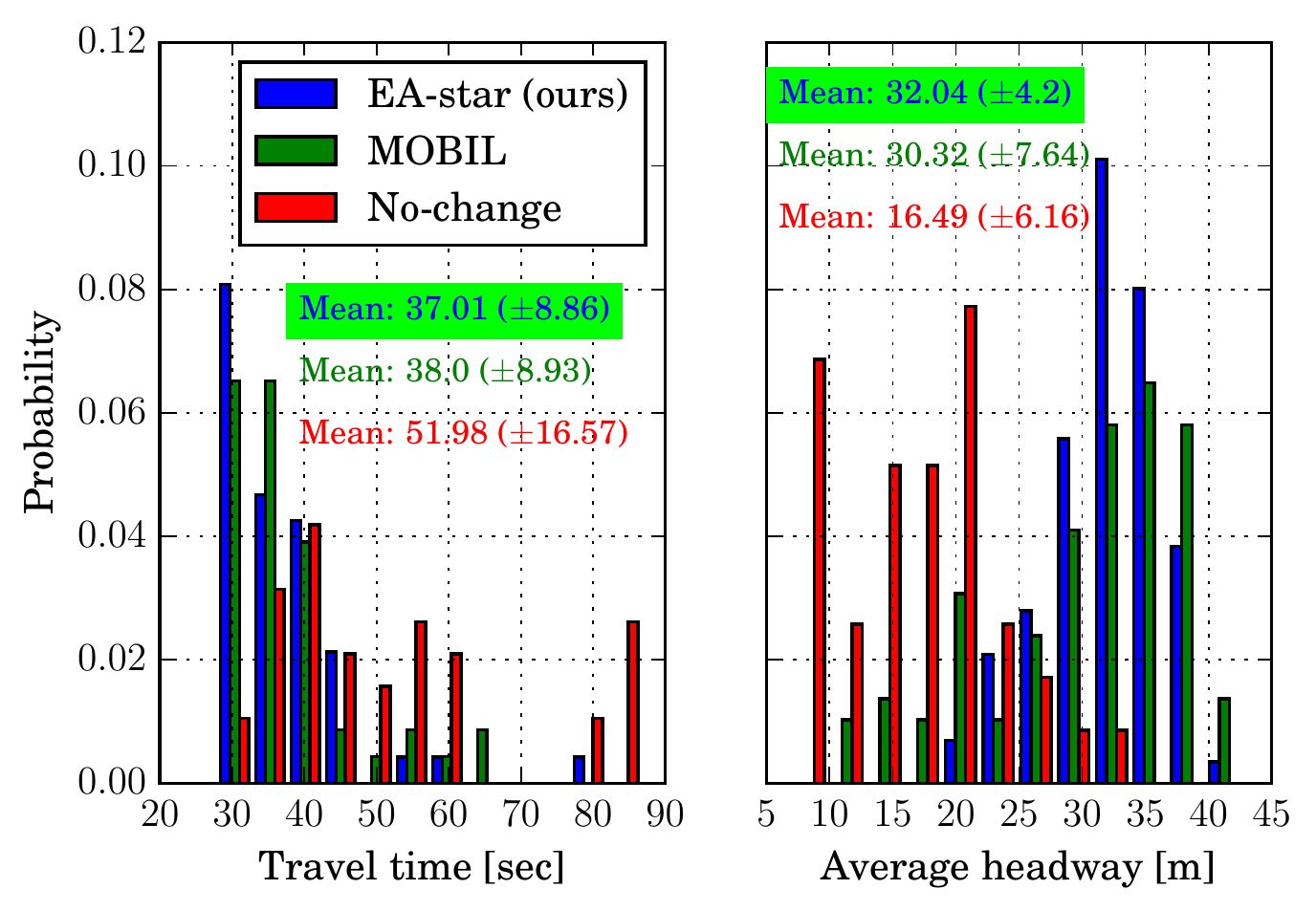}
    \caption{Monte Carlo simulation result for the travel time (left) and average headway (right). The numbers in parentheses show the standard deviation.}
    \label{fig:monte}
    \vspace{-10pt}
\end{figure}

Finally, Fig.~\ref{fig:monte} reports a quantitative analysis on the travel time and average headway among the three models. In each run of the Monte-Carlo simulations, other vehicles are initialized with random positions and velocities (thus, each run is a different scenario of driving on the highway). The results indicate that our model, EA$^\star$, is expected to outperform the other two models in terms of travel time and headway (visibility range), under multiple random cases. Also, our model secures a lowest standard deviation, implying the consistency in its performance. 

%
 




\section{CONCLUSIONS}\label{sec:conclusion}
\added[]{This work addresses a lane-selection problem with respect to uncertain and moving obstacles in relatively dense traffic.} Specifically, we propose a lane-selection algorithm extended to a widely applied path search algorithm, A$^\star$. The detailed cost configuration is developed, which comprehensively evaluates risk, travel time, and control efforts and guarantees solution existence. The proposed method is straightforward to implement while remaining computationally efficient ($\approx$0.005 [s] in each search). The performance of the proposed method is demonstrated under the Carla simulator, incorporating a state-of-the-art motion planning and control framework, Neural Network integrated Model Predictive Controller. A comparative analysis against a renowned lane selection model, MOBIL, shows that the proposed model outperforms in terms of travel time and headway. \added[]{Future works include dynamic speed profiles in the graph initialization and having a speed advisory system.}

\addtolength{\textheight}{-12cm}   




\section*{APPENDIX}\label{appendix}
\begin{prop}
Given a finite set of surrogate goals $G_s$, if the heuristic cost $h$ is admissible, A$^\star$ terminates by finding an optimal path (with an optimal choice of surrogate goal).
\end{prop}
\textit{Proof.} Let $g_1$ and $g_2$ the arbitrary surrogate goals in $G_s$. Suppose the path is optimal with $g_1$ but the algorithm terminates at $g_2$ with a suboptimal cost, i.e.
\begin{equation}
    f(g_1) < f(g_2)\label{eq:appendix_1}.
\end{equation} Then there must exist a node $n$ (in the open list) that
\renewcommand{\labelenumi}{\textbf{{[R\arabic{enumi}]}}}
\begin{enumerate}
\item is not expanded, i.e., $f(n) \geq f(g_2)$,
\item is an optimal path from start (as a part of the optimal path), i.e., $g(n) = g_\star(n)$, where $\star$ indicates optimal.
\end{enumerate}
Now, $f(n)$ reads
\begin{align}
    f(n) &= g(n) + h(n)\\
         &= g_\star(n) + h(n) \quad\text{by \textbf{[R2]}}\\
         &\leq g_\star(n) + h_\star(n) \quad\text{by admissibility}.
\end{align}
By \textbf{[R1]}, the following suffices
\begin{align}
    g_\star(n)+h_\star(n)&\geq f(n) \geq f(g_2)\\
    \rightarrow f(g_1) \geq f(g_2)\label{eq:appendix_2}.
\end{align}
Equation~\eqref{eq:appendix_2} contradicts \eqref{eq:appendix_1}, thus the proof concludes. Alternatively, the problem with surrogate goals can be transformed to the canonical problem (where one true goal exists) by adding a pseudo goal (with zero step and heuristic cost) that can be transitioned from any surrogate goals. 
\mbox{}\hfill
$\blacksquare$



\bibliographystyle{IEEEtran}
\bibliography{ref.bib}

\end{document}